\documentclass[11pt,onecolumn]{article}         
\usepackage[top=3.2cm, bottom=3.0cm, left=2.3cm, right=2.3cm]{geometry}

\usepackage{bbm}
\usepackage{color}
\usepackage{graphicx}
\usepackage{cite}
\usepackage[cmex10]{amsmath}
\usepackage{amssymb}
\usepackage{subfigure}
\usepackage{rotating}
\usepackage{multirow}
\usepackage{algorithm}
\usepackage{algorithmic}
\usepackage[colorlinks=true,linkcolor=red,citecolor=blue]{hyperref}
\usepackage{footmisc}
\usepackage{indentfirst}
\usepackage{array}
\usepackage{flushend}
\usepackage{framed,multirow}
\begin{document}

\title{GeoSay: A Geometric Saliency for Extracting Buildings\\ in Remote Sensing Images\footnote{All results are available at \url{http://captain.whu.edu.cn/project/geosay.html}.
}}

{
\author{Gui-Song~Xia$^{1}$, Jin~Huang$^{1}$, Nan~Xue$^{1}$, Qikai Lu$^{1,2}$, Xiaoxiang Zhu$^3$
\\
$^1${\em State Key Lab. LIESMARS, Wuhan University, Wuhan, 430079, China}\\
$^2${\em Electronic Information School, Wuhan University, Wuhan, 430079, China}\\
$^3${\em Civil, Geo and Environmental Engineering, Technische Universität München, Germany}\\
}
}

\maketitle

\begin{abstract}
Automatic extraction of buildings in remote sensing images is an important but challenging task and finds many applications in  different fields such as urban planning, navigation and so on.
This paper addresses the problem of buildings extraction in very high-spatial-resolution (VHSR) remote sensing (RS) images, whose spatial resolution is often up to half meters and provides rich information about buildings.  
Based on the observation that buildings in VHSR-RS images are always more distinguishable in geometry than in texture or spectral domain, this paper proposes a geometric building index (GBI) for accurate building extraction, by computing the geometric saliency from VHSR-RS images.
More precisely, given an image, the geometric saliency is derived from a mid-level geometric representations based on meaningful junctions that can locally describe geometrical structures of images.
The resulting GBI is finally measured by integrating the derived geometric saliency of buildings.
Experiments on three public and commonly used datasets demonstrate that the proposed GBI achieves the state-of-the-art performance and shows impressive generalization capability.
Additionally, GBI preserves both the exact position and accurate shape of single buildings compared to existing methods.
\end{abstract}
\definecolor{car}{RGB}{0,0,128}
\definecolor{parking lot}{RGB}{255,20,147}
\definecolor{plane}{RGB}{255,0,255}
\definecolor{baseball diamond}{RGB}{250,235,215}
\definecolor{bridge}{RGB}{0,128,0}
\definecolor{ground track field}{RGB}{127,255,212}
\definecolor{ship}{RGB}{165,42,42}
\definecolor{tennis court}{RGB}{138,43,226}
\definecolor{basketball court}{RGB}{222,184,135}
\definecolor{storage tank}{RGB}{95,158,160}
\definecolor{soccer ball field}{RGB}{127,255,0}
\definecolor{turntable}{RGB}{210,105,30}
\definecolor{harbor}{RGB}{255,255,0}
\definecolor{electric pole}{RGB}{100,149,237}
\definecolor{swimming pool}{RGB}{220,20,60}
\definecolor{lake}{RGB}{255,160,122}
\definecolor{helicopter}{RGB}{0,0,139}
\definecolor{airport}{RGB}{0,139,139}
\definecolor{overpass}{RGB}{184,134,11}
\definecolor{viaduct}{RGB}{184,134,11}
\definecolor{yellow}{RGB}{255,255,0}

\section{Introduction}
\label{sec:intro}

\begin{figure*}[t!]
	\centering
	\includegraphics[width = 0.9\linewidth]{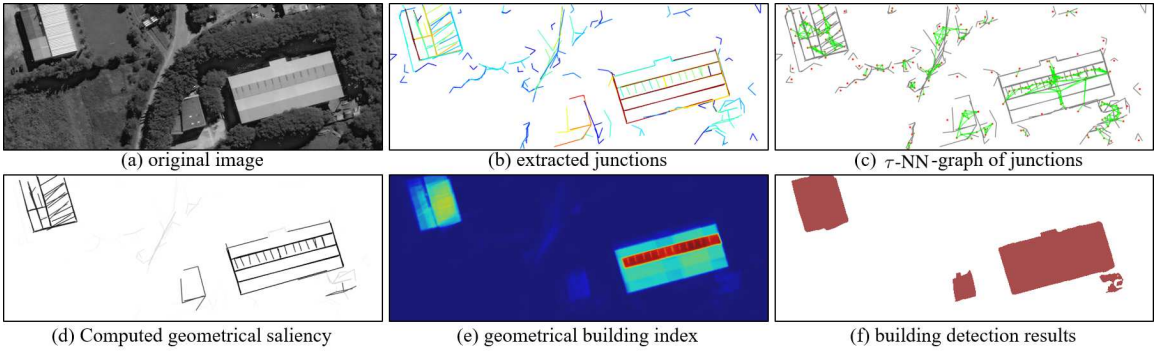} 
	\caption{Outline of the proposed method. Given an input image, we firstly detect the junctions by using ASJ detector. For each single junction, its reliability (NFA) and angle constraint from statistics contribute to its own building index. The neighboring junctions are found by K Nearest Neighbor algorithm with a distance constraint. The weight will be calculated based on the distance computed before. After combing the information of single junction and its neighbor, geometric building index will be computed for whole image. GBI will be blurred by a Gaussian kernel and shadow information will be added by applying black top-hat transform.}
	\label{fig:junc_detail}
\end{figure*}

Obtaining accurate locations and footprint shapes of buildings is an important task in remote sensing applications, and the generated building maps can be used in many fields, like urban mapping and planning, autonomous driving and so on, as presented by~\cite{Ghanea2016,Duan2004}.
In real applications, such perfect building maps are often achieved by manual administrations, which is laborious and expensive. 
As a result, the speed of updating building maps can not keep up with the pace of urbanization, especially in the cities that develop rapidly, e.g. most of the cities in China.
Nowadays, very high-spatial resolution (VHR) remote sensing (RS) images with spatial resolutions up to half meters, either from aerial or satellite platforms, can provide rich details of buildings and becomes popular data source for building mapping. 
Therefore it is highly demanded to develop some automatic methods for accurately extracting the locations and footprint shapes of buildings from VHR-RS images, e.g. see~\cite{Sirmacek2011,pesaresi_robust_2008,xu2015}.

In the past decades, many research have been dedicated to 
extract built-up areas and buildings from remote sensing images, e.g. see~\cite{martinez2005evaluating,pesaresi_robust_2008,sirmacek_urban_2009,huang_multidirectional_2011,liu2013perception,shao_basi:_2014}.
Among them, it is popular to develop some algorithms to compute building indexes in RS images. 
For instance, \cite{zha_use_2003-1} proposed the {\em normalized difference built-up index} (NDBI) to extract buildings in Landsat-TM images, by making use of the spectral features of buildings in the $4$-th and $5$-th bands of the multi-spectral images.
Beside spectral features, texture features have also been widely used for buildings extraction.
Based on the observation that pixels around buildings often have high local contrast because of shadows casting, \cite{pesaresi_robust_2008} proposed the {\em texture-derived built-up presence index} (Pantex), which utilized the texture information computed by gray-level co-occurrence matrix (GLCM) to extract built-in areas from satellite images.
As an extension, \cite{shao_basi:_2014} developed the {\em built-up areas saliency index} (BASI) by relying on multi-scale and multi-direction texture features measured with non-sampled Contourlet transform.
It demonstrated better results in built-in areas extraction than Pantex by~\cite{pesaresi_robust_2008}.
In contrast with texture features, geometrical and morphological profiles provide another aspect for extracting buildings in VHR-RS images. 
\cite{huang_multidirectional_2011} used multi-scale and multi-directional morphological operators to compute features of buildings in RS images and developed the so-called {\em morphological building index} (MBI). Actually, MBI has integrated several morphological characteristics of buildings such as brightness, shape, and size.

However, it is worth noticing that the geometric structures of buildings turn to be more and more important when the spatial resolution of RS images increase. Especially for VHR-RS images, it is often the prominent features of buildings.
For instance, \cite{martinez2005evaluating} have showed that the density of corners, e.g. Harris corners developed by~\cite{harris_combined_1988} or SUSAN corners proposed by~\cite{smith_susan_1995}), are efficient to distinguish man-made structures from natural objects.
\cite{sirmacek_urban_2009} have combined the SIFT key-points by \cite{lowe2004distinctive} with graph theory, and explored the relationships between local geometrical features.    
In contrast with other methods, it is more theoretical but with high computational complexity and time consuming.
Alternatively, \cite{sirmacek_urban_2010} later introduced another technique using Gabor feature points and spatial voting, which reported comparable results on same datasets but with much less time for building extraction.
Along this line, \cite{kovacs_improved_2013} developed a method by replacing Gabor filters with a new point feature detector, so-called Modified Harris for Edges and Corners (MHEC), and proposing an orientation-sensitive voting matrix.
More recently, \cite{liu2013perception} demonstrated that more clear geometrical profiles, such as precise corners and junctions detected by \cite{xia_accurate_2014}, was more efficient than local key=points for buildings extraction in VHR-RS images, and proposed the {\em perceptual building index} (PBI). 
Observe that PBI is robust to changes of image contrast and resolution variations due to the robustness of junctions detected.

In recent years, unlike above-mentioned methods that rely on hand-crafted features of buildings, approaches based on deep learning~\cite{Hu2015Transferring,deepzhuxiaoxiang} have been proposed to train end-to-end building detection models from a set of annotated images and turn to be one of the most popular directions.
\cite{saito_multiple_2016} employed convolutional neural networks (CNNs) as feature extractor to extract both buildings and roads simultaneously from aerial images.
In this methods, a five-layer multi-channel predicted CNN was designed, which took $64 \times 64$ image patches as input.
Meanwhile, special cost function and model averaging operations were used comparing with those introduced by~\cite{MnihThesis2013}.
More recently, \cite{zuo_hf-fcn:_2016} further improved the building extraction accuracy by developing a hierarchically fused fully convolutional network (HF-FCN).
It took the original image pixels as input and output the probability map of building category.
\cite{Shrestha2018} combined fully convolutional network and conditional random fields for building extraction, which reduced the noise (falsely classified buildings) and sharpened the boundaries of single buildings.
As we shall see in the experiments of our paper, those deep learning-based methods can achieved satisfied results on their training and testing dataset, but they often show limited generalization capability on images from different data sets and sensors.
In addition, collecting well-annotated training data is usually difficult and costs a lot of money and labors in real task.

This paper presents a new method for accurately detecting buildings in VHR-RS images, by computing the geometric saliency of buildings.
Our work is inspired by the observation that, in VHR-RS images, buildings are always more distinguishable in geometries (both local and global) than other features.
Instead of propagating probabilities through spatial voting like PBI, which will result in many redundant false pixels, we propose a geometric reasoning processing to extract the accurate position and shape of single buildings based on robust mid-level geometric representation.
More precisely, as illustrated in Fig.~\ref{fig:junc_detail}, we first propose to represent VHR-RS images with a mid-level geometrical representation, by exploiting junctions that can locally depict anisotropic geometrical structures of images.
We then derive the saliency of geometric structures on buildings, by considering the probability of each junction that measures its saliency to its surroundings and the relationship of junctions.
This process can encode both local and semi-global geometric saliency of buildings in images.
Finally, the geometric building index (GBI) of whole image is measured via integrating the computed geometric saliency (GeoSay).
A preliminary version of this work is presented by~\cite{igarss2018}.

In contrast with existing building indexes, our method results in less redundant non-building areas and can provide accurate location and geometric shape (contours) of buildings.
As we shall see in Section \ref{sec:experiment}, our method achieves the state-of-the-art performance\footnote{All results in this paper are available at \url{http://captain.whu.edu.cn/project/geosay.html}.} on both three public datasets.
Meanwhile GBI generates reasonable good results independently of satellites, scene categories or image contrast, and it shows promising generalization power to different datasets, especially in comparison with learning-based approaches.

The rest of this paper is organized as follows. Sec.~\ref{sec:preliminary} introduces the junctions and mid-level geometric structural representation of images.  Sec.~\ref{sec:statistics} analyzes the relationship between buildings and junctions. Based on the relationship, Sec.~\ref{sec:saliency} explains the details to compute geometrical building index from junctions. In Sec.~\ref{sec:experiment}, we will compare the proposed method with several state-of-the-art methods with three public and commonly used datasets. Finally, we draw some conclusion remarks in Sec.~\ref{sec:conclusion}.

\section{Preliminary: a junction-based representation of images}
\label{sec:preliminary}
As mentioned before, local geometric features, such as Harris corners by~\cite{harris_combined_1988}, have been employed to detect buildings in VHR-RS images for long years.
However, the inference of the exact position and shape of a building directly by local geometric features is still difficult.
Most of recent investigations can only find built-in areas rather than single buildings based on local geometric features.
In order to inference single buildings, we need to figure out whether the local structures like points and lines are located around buildings or not and then integrate related neighboring structures to a whole building.
While one of the main difficulties on this task is how to find the relationship between buildings and local geometric structures.

In this paper, we propose to explore the relationships between buildings and a specific local geometric structures, {\em i.e.} \emph{junction}.
As given by~\cite{xia_accurate_2014}, a \emph{junction} is defined as a local geometrical structure where several edges intersect together. Thus, a junction is composed of a central point (\emph{corner}) and several branches (\emph{edges}). In contrast with corners, such as Harris, junction is a mid-level geometric structure. 
Moreover, T-junctions and L-junctions are often distinguishable geometric features for man-made objects, e.g. buildings in remote sensing images. 

The detection of junctions has been studied for years but a detail review of them is out of the scope in this paper.
Here, we briefly describe the ASJ junction detector by~\cite{xue2017anisotropic}, that will be used in our work.

\begin{figure}[!t]
	\centering
	\includegraphics[width =  0.7\linewidth]{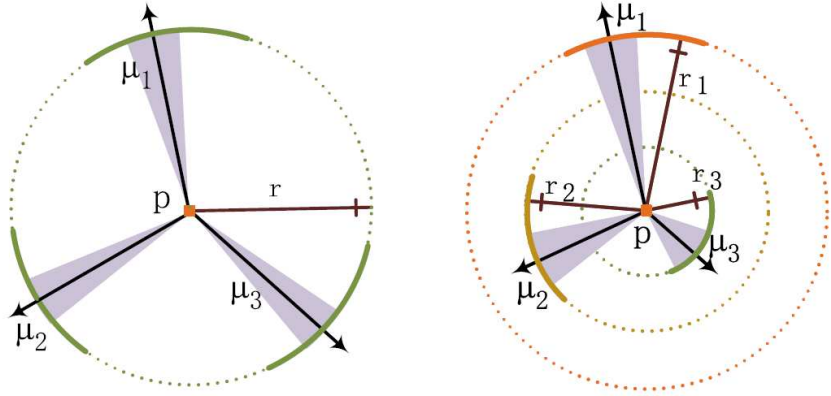}
	\caption{Template of isotropic-scale junction (left) defined in ACJ and anisotropic-scale junction (ASJ) (right), taken from \cite{xue2017anisotropic}. Junctions detected by ASJ have its own scale for each branch. }
	\label{fig:asj-junction}
\end{figure}

Consider a discrete panchromatic remote sensing image $U$ as a function $U : \Omega \to \mathbb{R}$, where $\Omega$ is an image grid.
A junction, illustrated by the right of  Fig.\ref{fig:asj-junction}, can be defined as,
\begin{equation}
\jmath:=\{\mathbf{p},\{s_i\}_{i=1}^M,\{\theta_i\}_{i=1}^M,\rho\},
\label{eq:junction-description}
\end{equation}
where $\mathbf{p} =(x,y)\in \Omega$ is junction's position in the image $U$, $M$ denotes the number of branches, $s_i \in \mathbb{R}$ and $\theta_i  \in [0, 2\pi)$ are the scale and orientation of the $i$-th branch respectively. $\rho$ is the significance of junction, and the smaller is better.

Detecting junctions in an image is to find all the local structures $\jmath$, modeled by the template illustrated on the right of Fig.\ref{fig:asj-junction}, and estimate their parameters. 
\cite{xia_accurate_2014} proposed the {a-contrario} junction detector (ACJ) with the help of the \emph{a-contrario} methodology~\cite{Desolneux2000}, where they assumed that the scale of branches are identical, {\em ,i.e.} $s_i=s_2=\ldots =s_M=s$. 
With ACJ detector, from the intersected point of a junction, we need to define a measurement to judge whether there exists junction or not and find those branches.
Each branch of a junction corresponds to an edge, thus the gradient inside branch's neighbors should be consistent with the direction of the branch.
Given a scale $s$, the neighbors of a branch with direction $\theta$ are defined as pixels inside a small sector $\mathbf{S_p}(s, \theta)$ along $\theta$ with radius $s$.
\begin{equation}
\mathbf{S_p}(s, \theta) := \Big\{\mathbf{q}\in \Omega; \mathbf{q}\ne \mathbf{p}, |\overrightarrow{\mathbf{p}\mathbf{q}}|\le s, d_{2\pi} (\alpha(\overrightarrow{\mathbf{p}\mathbf{q}}), \theta)\le \delta(s) \Big\},
\end{equation}
where $\delta(s)$ is a predefined parameter related to $s$, $\alpha(\overrightarrow{\mathbf{p}\mathbf{q}})$ is the angle of vector $\overrightarrow{\mathbf{p}\mathbf{q}}$ in [0, 2π] and $d_{2\pi}$ is the distance along the unit circle, defined as $d_{2\pi}(\alpha, \beta) = min(|\alpha-\beta|, 2\pi -|\alpha-\beta|)$.

When a branch corresponds to an edge, then most of the neighboring pixels should have similar direction of gradient with the direction of this branch.
Thus, the strength of a branch is measured inside its neighboring sector based on this idea.
For a given sector $\mathbf{S_p}(s, \theta)$, its strength is measured by
\begin{equation}
\mathbf{\omega_p}(s, \theta) = \sum_{\mathbf{q}\in \mathbf{S_p}(s, \theta)}{\gamma_p(\mathbf{q})},
\end{equation}
where $\gamma_p(\mathbf{q})$ is the pairwise strength between $\mathbf{p}$ and $\mathbf{q}$,
\begin{equation}
\begin{split}
\gamma_p(\mathbf{q}) = \vert\vert\nabla \tilde{\mathbf{U}}(\mathbf{q}) \vert\vert \cdot \max(|\cos(\phi(\mathbf{q}))-\alpha(\overrightarrow{\mathbf{p}\mathbf{q}})| \\
- |\sin(\phi(\mathbf{q}))-\alpha(\overrightarrow{\mathbf{p}\mathbf{q}})|, 0),
\end{split}
\end{equation}
where $\vert\vert\nabla \tilde{\mathbf{U}}(\mathbf{q}) \vert\vert$ is the normalized gradient of image $U$ in position $\mathbf{q}$, and $\phi(\mathbf{q})$ is the direction of the gradient.

With the strength of all branches, the strength $\mathbf{t}(\jmath)$ of junction $\mathbb{\jmath}$ is defined as the minimal strength of its $M$ branches.
But here we still need to set a parameter $t$ to find meaningful junction $\jmath'$ with enough larger strength, which means that $\mathbf{t}(\jmath') > t$.
\cite{xia_accurate_2014} has shown that one can compute a value called \emph{number of false alarm} ($\mathbf{NFA}$) based on the \emph{a contrario} theory to measure strength of a junction without parameter.
Meaningful junction should have $\mathbf{NFA}$ ranges from [0, 1], and the smaller the better.
In the definition of ACJ, $\rho$ corresponds to the value of $\mathbf{NFA}$.

Although ACJ is developed to detect junctions from nature images, there is no problem to deal with VHR remote sensing images.
However, the junctions detected by ACJ have only one scale, which means that the lengths of all branches are the same.
While buildings are often rectangular objects with unequal length of edges.
To deal with this problem, \cite{xue2017anisotropic} introduced an improved version of ACJ, called anisotropic-scale junction (ASJ) detector.
Based on the junctions detected by ACJ, ASJ can get anisotropic scales in various directions of junction's branches (see Fig.\ref{fig:asj-junction}, taken from \cite{xue2017anisotropic}). More details on ACJ and ASJ detectors can be found in the work of \cite{xia_accurate_2014} and \cite{xue2017anisotropic}.

In this paper, we employ the state-of-the-art ASJ detector to extract junctions to build a mid-level geometric profiles of buildings. Thus, given a panchromatic remote sensing image $U$, we can represent it by a set of $K$ detected junctions 
$
J = \big\{ \jmath_k \big\}_{k=1}^K.
$

\section{Statistics of junctions in VHR RS images}
\label{sec:statistics}

\begin{figure}[t!]
	\centering
	\begin{minipage}[c]{0.47\textwidth} 
		\centering 
		\includegraphics[width = 0.9\linewidth]{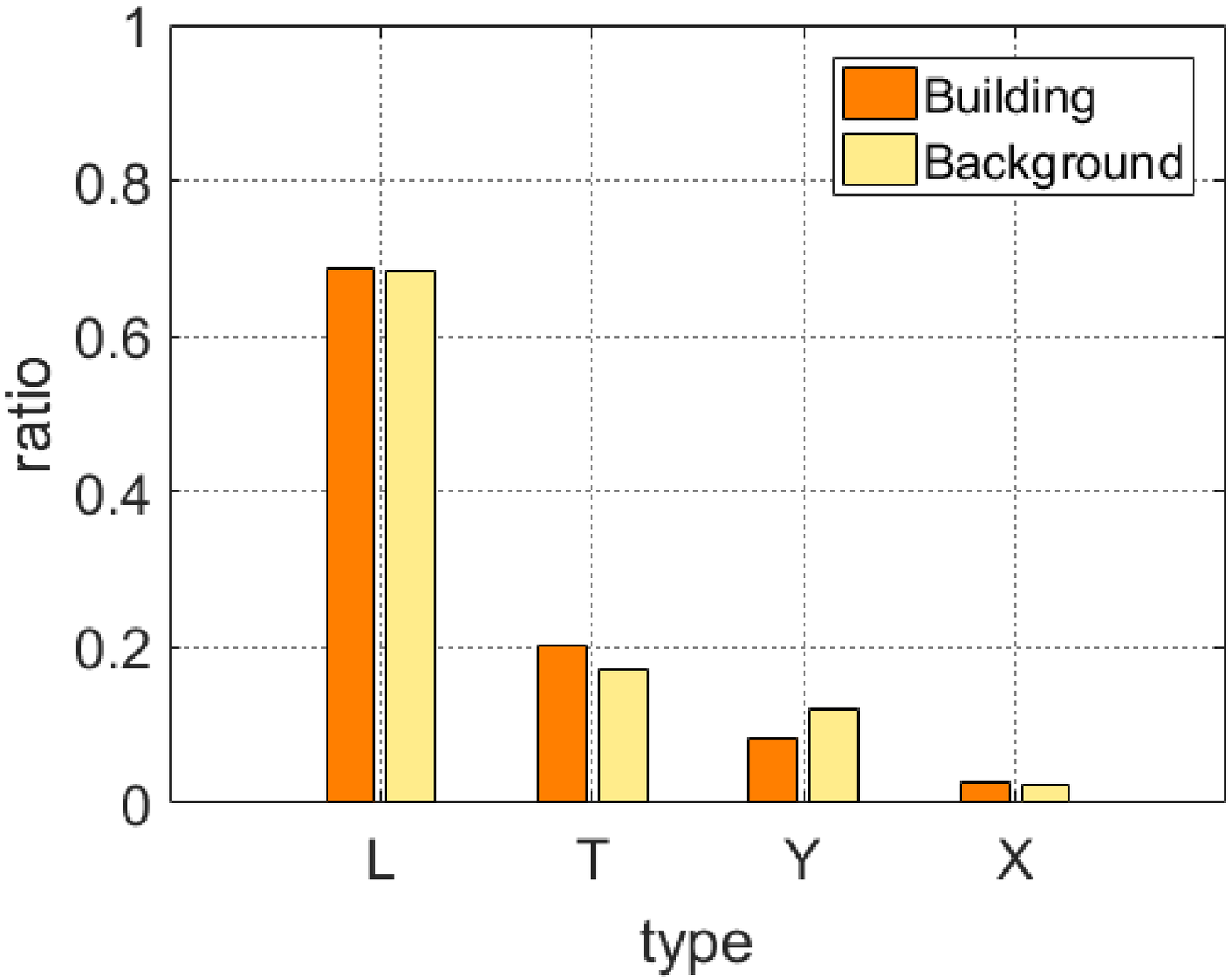} 
		\label{fig:statistic-types}
	\end{minipage}
	\begin{minipage}[c]{0.47\textwidth} 
		\centering 
		\includegraphics[width = 0.9\linewidth]{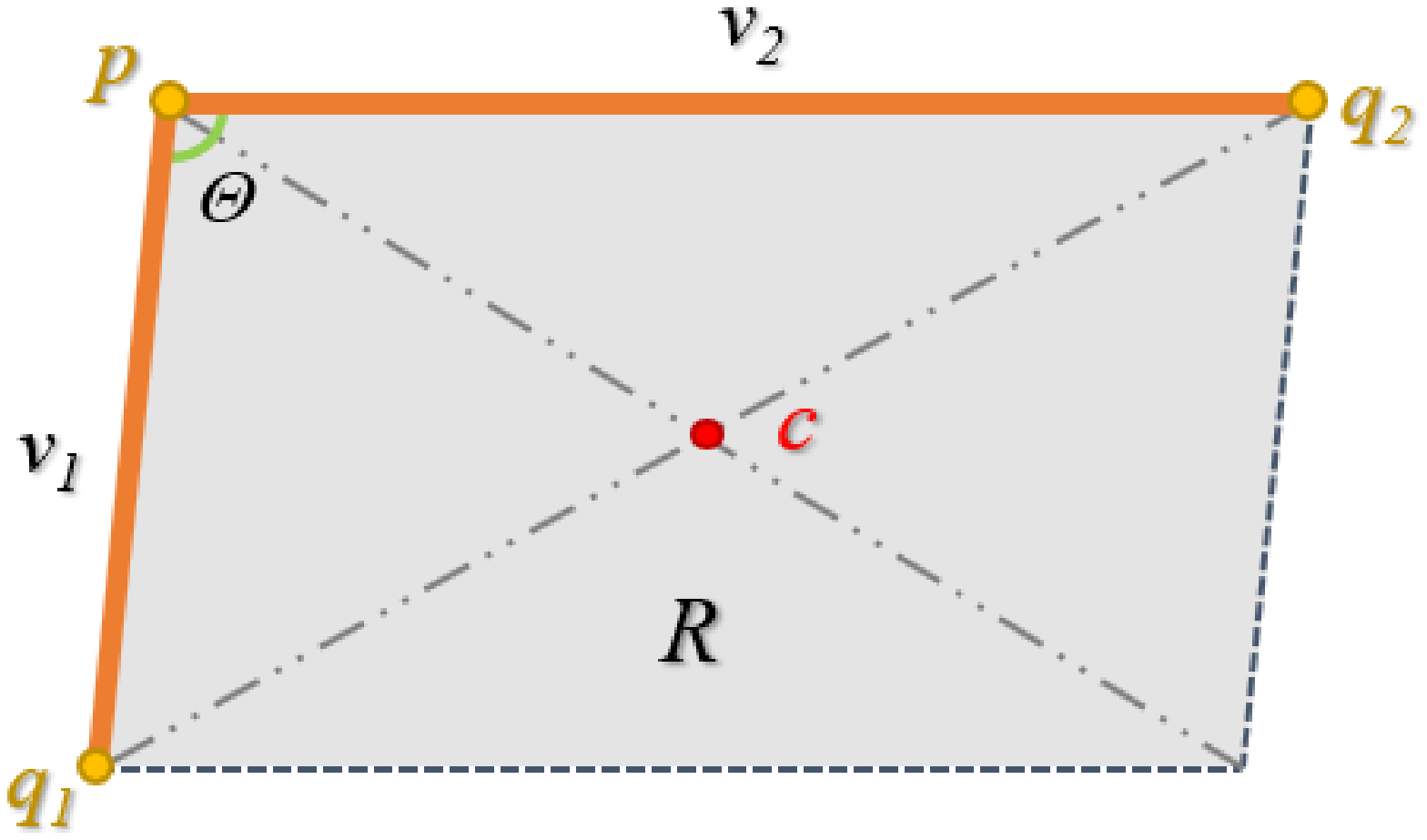} 
		\label{fig:junction-description}
	\end{minipage} 
	\caption{Left: The ratio of different types of junctions among building and background areas. Obviously, L-junction is the main type in both two areas. Right: The structure of $L$-junction. A $L$-junction contains a corner point $\mathbf{p}$, two branches ($\vec \nu_1,\vec \nu_2$) with their endpoints ($\mathbf{q}_1, \mathbf{q}_2$) and the included angle $\Theta$ of the two branches. Meanwhile, each junction could form a parallelogram region ($\mathbf{R}$) and the center $\mathbf{c}$ of a junction is defined as the center of its parallelogram.}
\end{figure}

Junctions can be divided to different types based on the number of branches. Such as $L$-junction has two branches ($M=2$), $T$/$Y$-junction and $X$-junction have three and four branches respectively. 
It is clear that $L$-junctions often correspond to object corners, while $T$-junctions imply occlusions between objects in images. $Y$-junctions and arrow-junctions usually correspond to corners of 3D objects. Junctions with order higher than $4$ are less discriminative. 
In order to verify this assumption, we counted the ratio of junctions with different types detected from the building areas and background areas in the Spacenet65 dataset (details of datasets shall be described in Section \ref{sec:experiment}). 
To judge whether a junction $\jmath$ is located along buildings or not, we validate the overlapping ratio between the areas of buildings and the area of parallelogram region $R_{\jmath}$ spanned by the junction. If the area of overlapping region between $R_{\jmath}$ and ground truth is larger than $0.8$, the junction $\jmath$ is thought to be located in the building area.
We computed the distribution of junction types with $200k$ junctions collected from Spacenet65 dataset. The result is showed in Fig.\ref{fig:statistic-types}.
Observe that, in both building area and background area, the distribution of junction type are similar: L-junctions are the dominant while junctions with more than $4$ branches are rare.

It is also worth noticing that $L$-junction is a fundamental element and all junctions with any type could be represented by several L-junctions.
Considering the computation complexity of using different junction types and this observation, we finally decide to only use $L$-junction.
To fully utilize all junctions, junctions with more than 3 branches were separated into several $L$-junctions.
For convenience, we rewrite the definition of $L$-junction as below
\begin{equation}
\jmath : \{\mathbf{c}, \vec \nu_1, \vec \nu_2, \beta, \rho \}
\end{equation}
where $\vec \nu_1, \vec \nu_2$ are the two branches of $L$-junctions and $\vec \nu_i = \overrightarrow{\mathbf{p}\mathbf{q}_i}$, with $\mathbf{q}_i = \mathbf{p} + s_i \cdot (\cos \theta_i, \sin \theta_i)^T$ for $i=1,2$. $\mathbf{c} = \frac{(\mathbf{q}_1 + \mathbf{q}_2)}{2}$ is the center of the $L$-junction $\jmath$. $\beta = \min\{| \theta_1-\theta_2 |,2\pi - | \theta_1-\theta_2 |\}$ is the included angle of junction's two branches. The significance $\rho$ inherits from its original junctions.
The details of a $L$-junction is showed in Fig.\ref{fig:junction-description}.

\begin{figure}[t!]
	\centering
	\subfigure{
		\includegraphics[width = 0.45\linewidth]{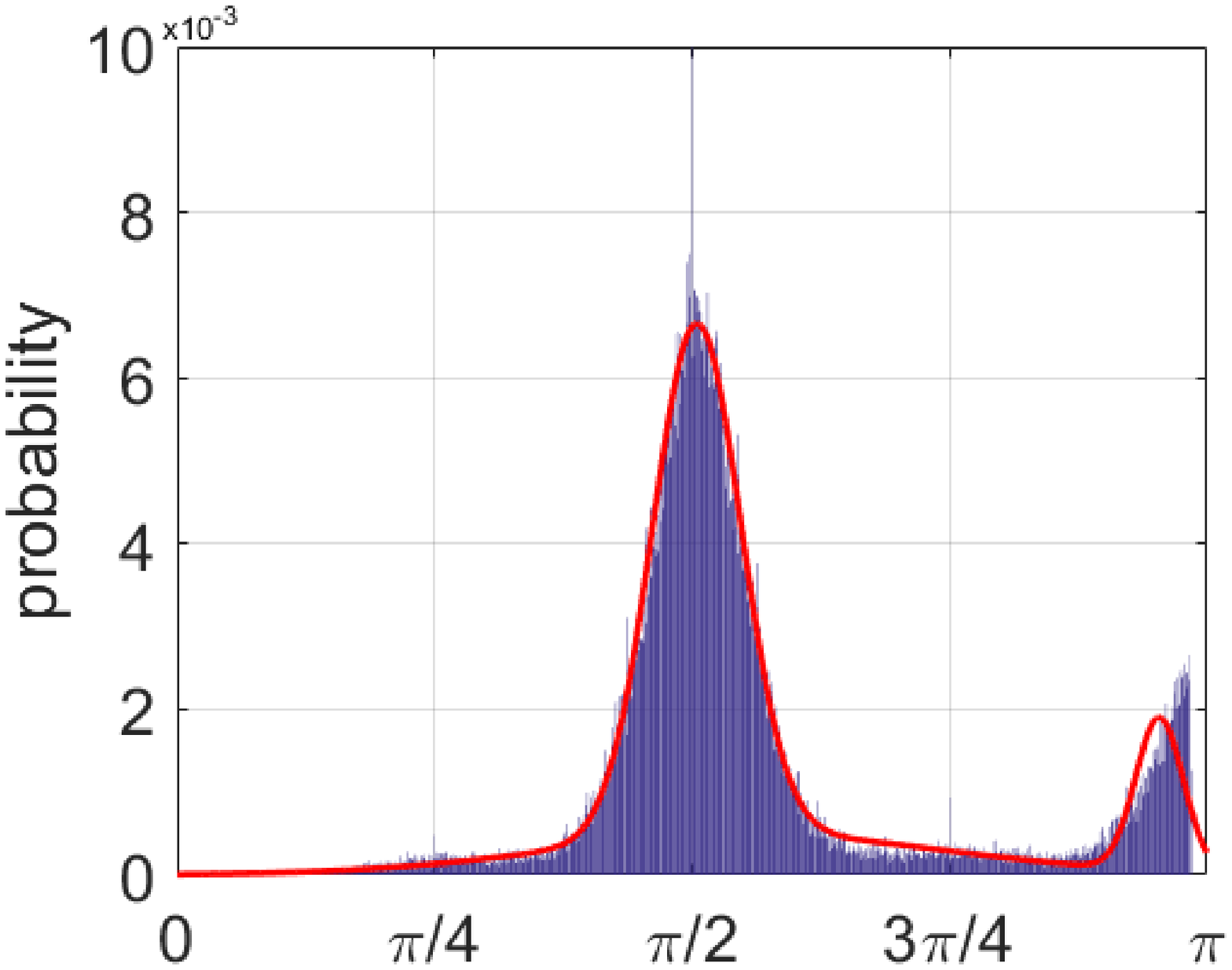}}
	\subfigure{
		\includegraphics[width = 0.45\linewidth]{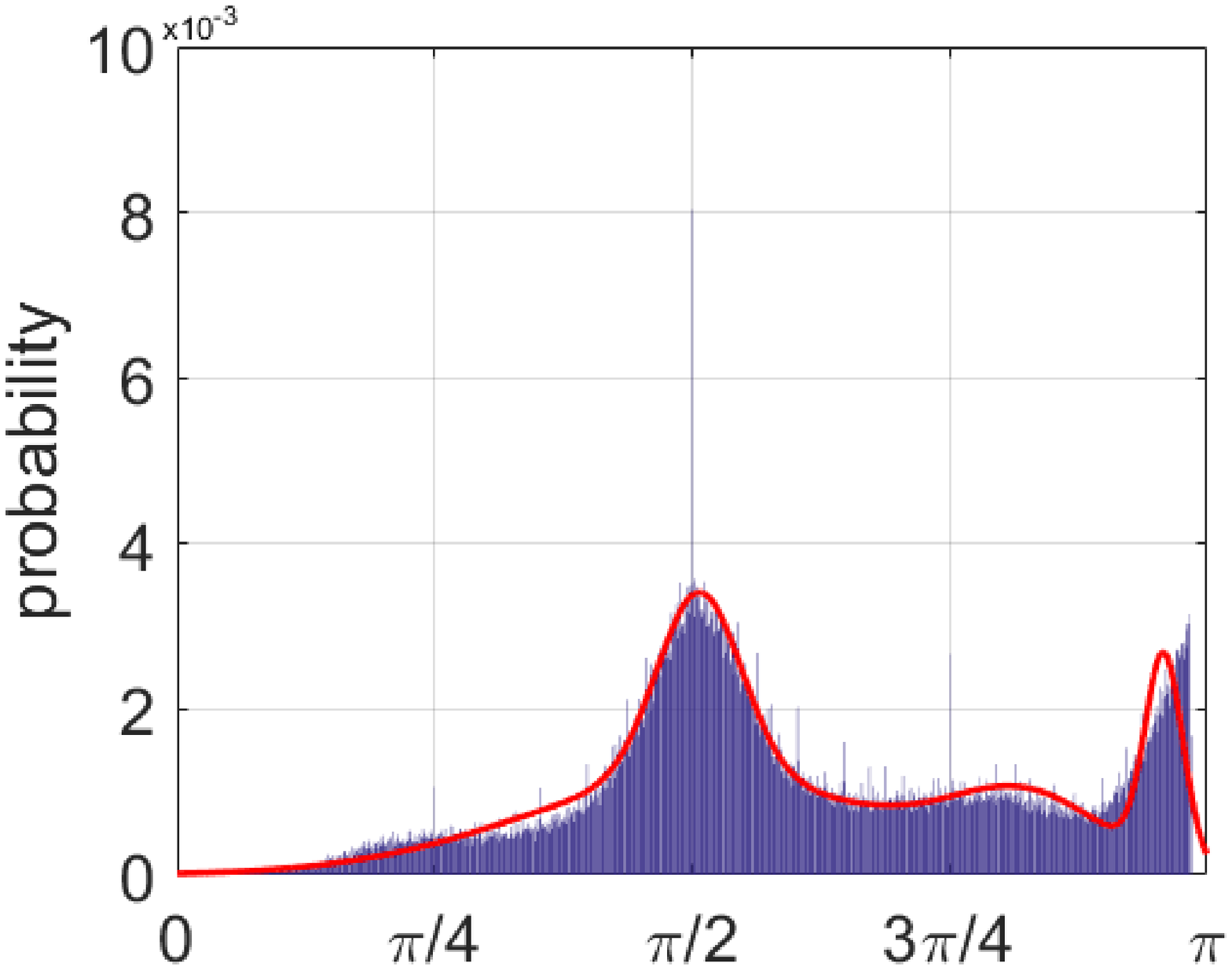}}
	\caption{The histogram in blue represents the distribution of L-junctions' included angles (x axis) among building area(left) and background area(right). The result showed that the angles around building are concentrated on $\pi/2$. The fitted curve is showed in red and we can see that the distribution is well simulated by it.}
	\label{fig:statistic-theta}
\end{figure}

Another observation is that, in RS images, the statistics of junctions should different on buildings and background. As junctions are detected along areas with high gradients, they are likely to be found around corners of buildings.
Buildings are typical man-made objects and their shape are usually very regular or more precisely, rectangular.
Thus, the included angles $\theta$ of L-junctions will also have special distribution when they are located along buildings.
To verify this supposition, we calculated the distributions of L-junctions' included angles among different regions from the Spacenet65 dataset, as illustrated in Fig.\ref{fig:statistic-theta}. One can find that junctions' included angles are really close to $\pi/2$ in the building area and have a large difference towards junctions among background area.
In building area, angles of junctions are highly concentrated in interval $[\frac{\pi}{3},\frac{2\pi}{3}]$.
While in the background area, distribution around $\frac{\pi}{2}$ has less contrast to other intervals.
Such distributions can help us to distinguish junctions around buildings from other objects, and can be used as a prior in the detection of buildings. 
In order to parameterize these distributions, we fit the two distributions by Gaussian Mixture Models (GMM, \cite{Mclachlan2000Finite}).
The distribution of angles and the fitted parametric probability curve are showed in Fig.\ref{fig:statistic-theta}.

In fact, we also counted other properties of junctions among different areas, like scale and position.
But the results show that the distributions of those attributes have little effect to distinguish buildings and backgrounds compared with angles.
\section{GeoSay: from junctions to building index}
\label{sec:saliency}

In this section, we will mainly explain the definition of geometric saliency (GeoSay) based on junctions and the details of proposed geometric building index (GBI).

\subsection{Geometric saliency}
\label{sec:geo-sa-building}

Buildings in VHR images, are often geometrically composed of several parallelograms (sometimes even rectangles).
A regular building may have several corners where L-junctions will be detected.
$L$-junction is likely to be detected at the corner of a building, and the junction's two branches will coincide with the two edges.
Thus the parallelogram spanned by the junction's branches will have a lot of overlapping parts with the buildings.
In such case, the spanned parallelogram could precisely represent a part of buildings and preserve the geometric shape.

Based on ASJ detector, we could have a mid-level geometric representation of the whole image.
And our task is to find salient junctions that are located at buildings' corners.
Once we find such junctions, buildings could be detected based on the relationship between junctions' parallelograms and buildings.
Here we defined first-order and pairwise geometric saliency to find salient junctions, based on the properties of single junctions and the relationship between neighboring junctions.

\subsubsection{First-order geometric saliency}

There are two important characteristics when junctions are located along buildings.
First, for an image $U$, the significance $\rho_\jmath$ of each junction $\jmath$ indicates the reliability of detection.
The smaller the $\rho$ is, the more reliable the junction will be.
Secondly, as we have shown in Section \ref{sec:statistics}, the distributions of included angles $\beta$ are statistically dsicriminative between buildings and backgrounds.

Given an image, all detected junctions $\mathcal{J}$ can be divided into two subsets, {\it i.e.,} $\mathcal{J} = \mathcal{J}_{B} \cup \mathcal{J}_{\bar B}$, $\mathcal{J}_B$ inside buildings and $\mathcal{J}_{\bar B}$ outside buildings.
For a junction $\jmath$ with its parametric description  $\Theta_\jmath = \{\mathbf{c}, \vec \nu_1, \vec \nu_2, \beta, \rho \}$, the posterior probability $\mathbb{P}(\jmath \in \mathcal{J}_B \, | \, \Theta_\jmath)$, measuring the possibility of the event that a junction $\jmath$ is inside buildings, can be derived by

\begin{equation}
\mathbb{P}(\jmath \in \mathcal{J}_B \, | \, \Theta_\jmath) = \frac{\mathbb{P}( \Theta_\jmath \,| \, \mathcal{J}_B ) \mathbb{P}(\mathcal{J}_B)}{\mathbb{P}( \Theta_\jmath \,| \, \mathcal{J}_B ) \mathbb{P}(\mathcal{J}_B) + \mathbb{P}( \Theta_\jmath \,| \, \mathcal{J}_{\bar B} ) \mathbb{P}(\mathcal{J}_{\bar B})}
\label{eq:jbs-jtheta}
\end{equation}
where the prior probabilities $P(\mathcal{J}_B), \, P(\mathcal{J}_{\bar B})$ and the likelihoods $\mathbb{P}( \Theta_\jmath \,| \, \mathcal{J}_B )$, $\mathbb{P}( \Theta_\jmath \,| \, \mathcal{J}_{\bar B})$ can be estimated from a given dataset of buildings, e.g. the Spacenet65 dataset, based on the fitted GMM model.

By combining the significance parameter and included angle of a single junction, the first-order geometric saliency $g_\jmath^{(1)}$ of a junction $\jmath$ can be computed as
\begin{align}
g_\jmath^{(1)} = (1- \rho_\jmath) \cdot \mathbb{P}(\jmath \in \mathcal{J}_B \, | \, \Theta_\jmath).
\label{eq:first-order}
\end{align}
which indicates the degree of a single junction locating along buildings.

\subsubsection{Pairwise geometric saliency}

First-order geometric saliency encodes the properties of single junctions, and pairwise geometric saliency is defined for utilizing the relationship between neighboring junctions.

\begin{figure}[t!]
	\centering
	\subfigure{
		\includegraphics[width = 0.4\linewidth]{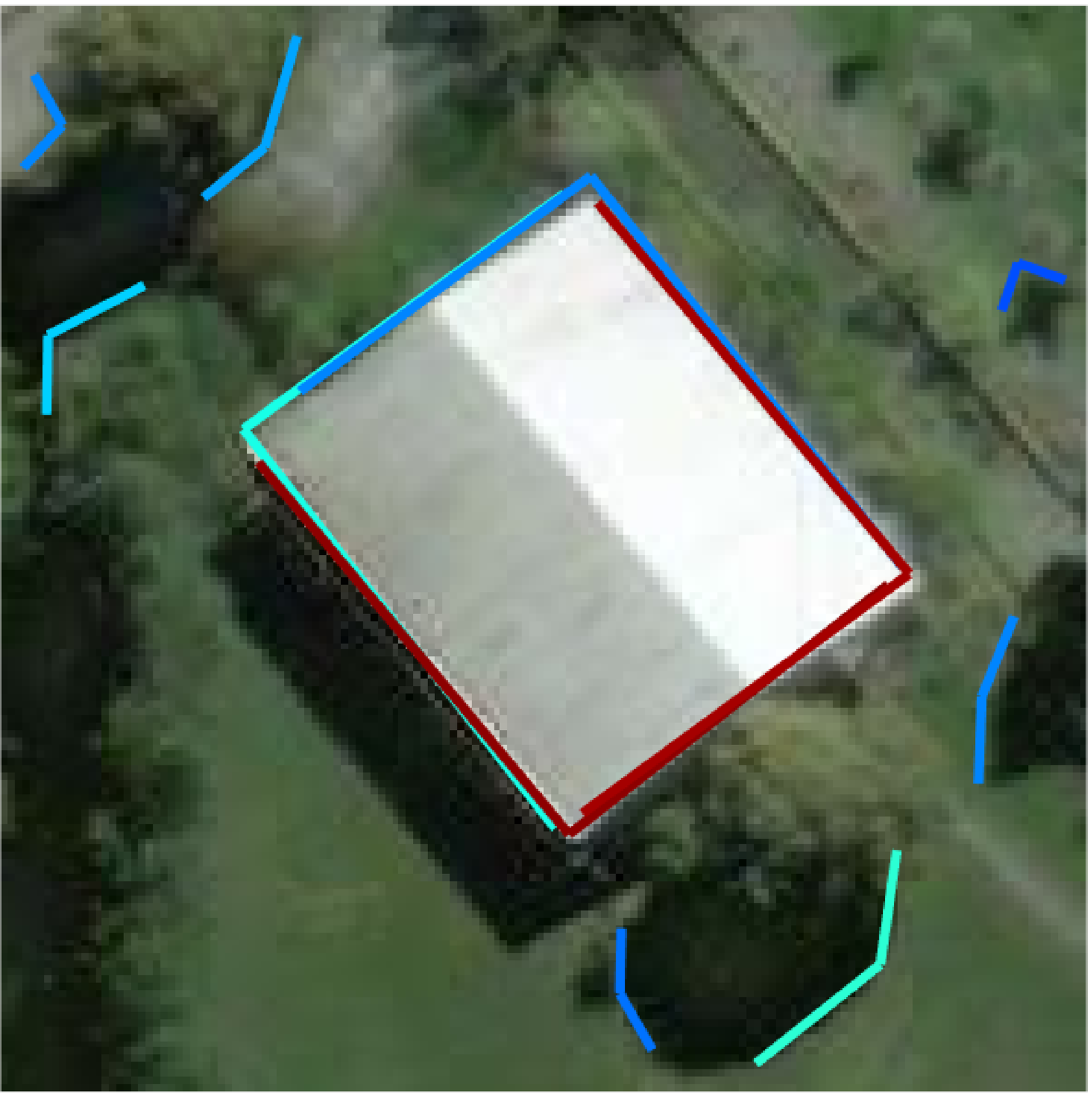}}
	\subfigure{
		\includegraphics[width = 0.4\linewidth]{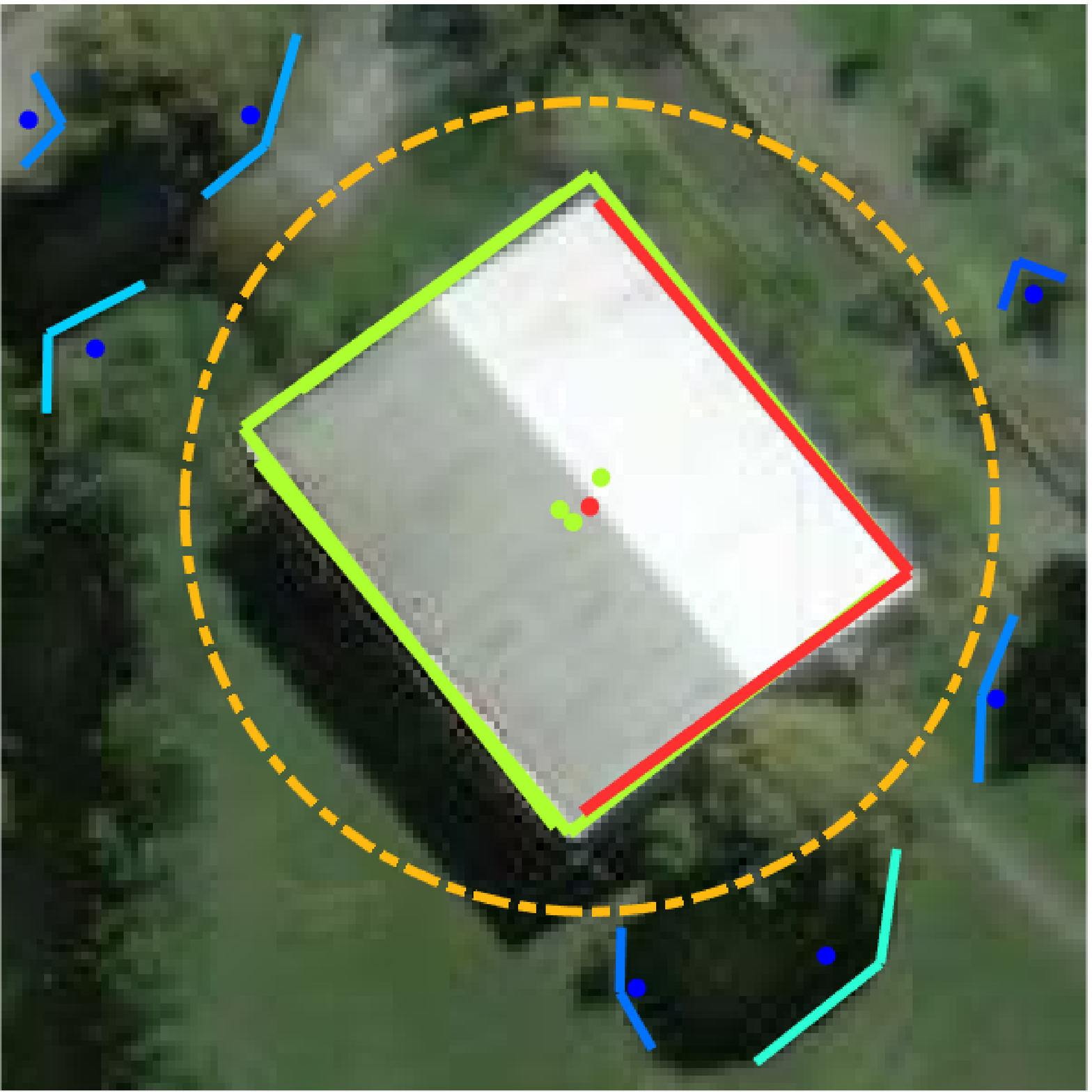}}
	\caption{Left: junctions detected by ASJ detector. Right: neighbors of the selected red junction. Its neighbors are those green junctions whose centers satisfied the Eq.\ref{eq:distance-constraint}.}
	\label{fig:junc_neighbor}
\end{figure}

When there are many junctions whose centers are very close to each other in a region, the probability of existing a building will be higher. Thus, pair-wise relationships of junctions are useful cues to derive geometric saliency. In contrast with first-order saliency, pair-wise ones can encode more globally geometric information in images.
Here, we use nearest neighbors to compute pair-wise saliency.
For a junction $\jmath$, its $\tau$-{\it nearest neighbors} ($\tau$-NN), denoted by $\mathcal{N}_\jmath$, is defined as a set of junctions satisfying
\begin{equation}
\| \mathbf{c}_\jmath - \mathbf{c}_{\jmath'} \|_2 < \tau, \, \forall \jmath' \in \mathcal{N}_\jmath.
\label{eq:distance-constraint}
\end{equation}
where $\tau$ represents the maximal length of branch of the junction $\jmath$. An example of junctions and neighboring junctions is displayed in Fig.\ref{fig:junc_neighbor}, where green points are the centers of junctions inside the $\tau$-NN of the junction with location center in red. 
Based on neighboring junctions, the pair-wise geometric saliency of a junction $\jmath$ is defined as,
\begin{equation}
g_\jmath^{(2)} = \sum_{\jmath' \in \mathcal{N}_\jmath} e^{-\tau^{-1} \cdot \| \mathbf{c}_\jmath - \mathbf{c}_{\jmath'} \|_2} \cdot g_{\jmath'}^{(1)}.
\label{eq:pairwise}
\end{equation}

Furthermore, neighboring junctions should have similar scales as they are located in the same buildings.
Besides the distance constraint, there will also be a scale constraint.
If the ratio of scales between junctions are too large, the neighboring junction will be discarded from the neighbor list.

\subsection{Geometric building index}
\label{sec:gbi}

Note that, given an $L$-junction $\jmath : \{\mathbf{c}, \vec \nu_1, \vec \nu_2, \rho \}$, the two branches $\vec \nu_1, \vec \nu_2$ uniquely span a parallelogram $R_\jmath$, as shown in Fig.~\ref{fig:junction-description}. 
Our {\it geometric building index} (GBI) attempts to associate each pixel $\mathbf{p}$ with a saliency measuring the possibility of the pixel belonging to buildings, which is the summation of saliency inside parallelogram of all junctions.
Thus, for a pixel $\mathbf{p} \in \Omega$ in $U$, we calculate its building index by:
\begin{equation}
\textrm{GBI}(\mathbf{p}) = \sum_{\jmath \in \mathcal{J}} \big( g_\jmath^{(1)} + g_\jmath^{(2)} \big) \cdot \,\mathbbm{1}_{\mathbf{p} \in R_\jmath}.
\label{eq:gbi}
\end{equation}
where $\mathcal{J}$ is the list of junctions detected by the ASJ detector in image $U$,
and $\mathbbm{1}_{\mathbf{p} \in R_\jmath}$ is an indicator function, which equals $1$ if the pixel $\mathbf{p}$ is inside the parallelogram $R_\jmath$ of junction $\jmath$ and equals to $0$ otherwise.

Furthermore, inside VHR-RS images, there are many shadows.
The shape of building's shadow is often regular and junctions will be detected there.
\cite{huang_morphological_2012} have applied the black top-hat transform to extract shadow.
We were inspired by this idea and apply black top-hat transform in the brightness channel of the original image.
The transformed image is calculated by applying morphological closing operation to the original image and using the result to subtract original image.
We use one subtract this transformed image as a new multiplicative suppression term to the computation of GBI.

\subsection{Numerical implementation of geometric building index}
The code of ASJ detector was provided by the author in github\footnote{The code of ASJ could be found in \url{https://github.com/cherubicXN}}, written in C++.
The algorithm of calculating geometric building index was implemented using Matlab and the pseudocode is showed in Algorithm \ref{alg:gbi}.

\begin{algorithm} [htb!]
	\caption{Computation of geometric building index}
	\label{alg:gbi}
	
	\begin{algorithmic}
		\STATE \textbf{Input:} Image $U$;
		\STATE \textbf{Output:} matrix $G$ containing geometric building index;
		\STATE
		\STATE \textbf{\em //** Step 1: junction detection **//}
		\STATE $\mathcal{J} \gets $ Apply ASJ detector on $U$;
		\FORALL{$\jmath \in \mathcal{J}$}
		\STATE $\mathbf{q}^\jmath_i = \mathbf{p}_\jmath + s^\jmath_i \dot (\cos\theta^\jmath_i, \sin\theta^\jmath_i )$
		\STATE $\mathbf{c}_\jmath = \frac{(\mathbf{q}^\jmath_1 + \mathbf{q}^\jmath_2)}{2}$
		\STATE $\Theta_\jmath = \min\{| \theta^\jmath_1-\theta^\jmath_2 |,2\pi - | \theta^\jmath_1-\theta^\jmath_2 |\}$
		\STATE $N(\jmath) \gets$ the neighbors satisfied the constraint in Eq.\ref{eq:distance-constraint};
		\ENDFOR
		
		\STATE
		\STATE $[m,n] = size(U)$;
		\STATE $G = zeros(m,n)$;
		\FORALL{$\jmath \in \mathcal{J}$}
		\STATE \textbf{\em //**Step 2: first-order geometric saliency **//}
		\STATE $g_\jmath^{(1)} = (1- \rho_\jmath) \cdot \mathbb{P}(\jmath \in \mathcal{J}_B \, | \, \Theta_\jmath)$;
		\STATE $g_\jmath^{(2)} = 0$;
		\STATE $\tau = max(s^\jmath_1, s^\jmath_2)$;
		\STATE
		\STATE \textbf{\em //** Step 3: pair-wise geometric saliency **//}
		\FORALL{$\jmath' \in N(\jmath)$}
		\STATE $ga_\jmath'^{(1)} = (1- \rho_\jmath') \cdot \mathbb{P}(\jmath' \in \mathcal{J}_B \, | \, \Theta_\jmath')$;
		\STATE $g_\jmath^{(2)} = g_\jmath^{(2)} + e^{-\tau^{-2} \cdot \| \mathbf{c}_\jmath - \mathbf{c}_{\jmath'} \|_2^2} \cdot g_{\jmath'}^{(1)}$;
		\ENDFOR
		\STATE
		\STATE \textbf{\em //** Step 4: geometric building index **//}
		\STATE $R_\jmath \gets $ parallelogram bounded by $\jmath$;
		\FORALL{$\mathbf{p} \in R_\jmath$}
		\STATE $G(\mathbf{p}) = G(\mathbf{p}) + \omega_\jmath^{(1)} + \omega_\jmath^{(2)}$;
		\ENDFOR
		\ENDFOR
		
		\STATE
		\STATE \textbf{\em // **Step 5: plus shadow information **//}
		\STATE $U' \gets $ Apply Black Top-hat transform in $U$
		\STATE $G = G.*(1-U')$
	\end{algorithmic}
\end{algorithm}

The whole algorithm could be divided in to five steps. Step 1 detects junctions from the input, VHR-RS images, and computes the parameters of junctions ahead. Step 2 computes first-order geometric saliency of every junction based on its significance and information of angles. By using neighboring junctions, Step 3 generates pair-wise geometric saliency. Then Step 4 generates building index for each pixel based on the relationship between junction's parallelograms and buildings. Step 5 integrates shadow information into building index by using black top-hat transform.
\vspace{-3mm}
\section{Experiments and analysis}
\label{sec:experiment}

In this section, we evaluate the proposed method, {\em i.e.} GBI, and compare it with the state-of-the-art on three public and commonly used datasets. 
The involved building extraction algorithms includes BASI (\cite{shao_basi:_2014}), MBI (\cite{huang_multidirectional_2011}), PBI (\cite{liu2013perception}), which do not need any training data.
We also compared it with a deep learning-based method, {\em i.e.} HF-FCN (\cite{zuo_hf-fcn:_2016}), for which we used the model provided by the authors in github\footnote{The model of HF-FCN is provide by the authors and can be found at \url{https://github.com/tczuo/HF-FCN-for-Robust-Building-Extraction}}.
We also make some ablation studies on the GBI. 

\subsection{Experimental setups}
\subsubsection{Datasets}
We used three public datasets as follows.
\begin{itemize}
	\item[-] {\bf Spacenet65 dataset by~\cite{spacenetamazon}}:.
	SpaceNet is a corpus of commercial satellite imagery and part labeled data, available for academic usage.
	Our experimental data comes from the Area of Interest 1 (AOI 1) at Rio de Janeiro.
	This dataset collected imagery from DigitalGlobe WorldView-2 satellite with spatial resolution $0.5$m.
	We used the RGB imageries with labeled building footprints.
	The original size of satellite images are $19584 \times 19584$ pixels.
	Considering the feasibility of testing algorithms, we cropped the original data to a bunch of smaller images with fixed size as 2000$\times$2000 pixels.
	Note that not all cropped images contain buildings and also the given building footprints are not completed.
	Therefore we only pick cropped images with buildings and whole building footprints marked.
	After that, we manually correct every building footprint for chosen images.
	The final Spacenet65 dataset includes 65 images of 2000$\times$2000 pixels and their ground truth.
	This dataset covers mostly urban and rural areas and buildings of different appearances.
	
	\item[-] {\bf Massachusetts dataset} is a dataset designed for training neural network for building detections, proposed by \cite{MnihThesis2013}.
	It contains three subsets, 131 images for training, 4 images for validation and 10 images for test.
	The images are 1500$\times$1500 pixels with resolution of 1m.
	Due to the disordered gradients of images (difficult to extract local structures), we smoothed all images with a small 3x3 Gaussian kernel while applying ASJ detector.
	
	\item[-] {\bf Potsdam dataset}: 
	It is published by \cite{potsdam} and contains 38 patches (of the same size), each consisting of a true orthophoto (TOP) extracted from a larger TOP mosaic.
	Here, we only used the ortho corrected images.
	They are all very large images with size 6000$\times$6000 and we also cropped them like before to 2000$\times$2000.
	After deleting images without buildings, we finally get 214 images with their ground truth.
	The resolution of ground truth is 0.05m.
	Due to the high resolution, buildings inside this dataset are a little bigger than others.
\end{itemize}

\subsubsection{Settings of parameters}
The distributions of junctions' included angles among different areas are fitted from Spacenet-65 dataset by applying Gaussian Mixture Model (GMM) in the collected over 200k junctions.
Based on the individual shape of the two distributions among building and background areas, they are fitted respectively by 3 and 4 Gaussian models.

In general, a rectangular building could be fully represented by 4 L-junctions.
Thus, here we set $\tau = 4$ in the processing of finding neighbors.
The ratio of scales between neighboring junctions is set to 3.
It means that when the scale of a neighbor junction $\jmath'$ is three times larger or smaller than the junction $\jmath$, $\jmath'$ will be discarded.

While blurring, the size of the Gaussian kernel is set to 5-by-5, while $sigma$ is set to 0.5.
The kernel of black top-hat transform is selected as a $50 \times 50$ square.

\subsubsection{Evaluation metric}
To evaluate the accuracy of detection, we employed two commonly used metrics: mean Average Precision(\textit{mAP}) \cite{Buckley_2000} and \textit{F-score} \cite{powers2011evaluation}.
As directly assessing generated index map is difficult, thus we used thresholds ranges from [0,1] with step ${\Delta}r = 0.01$ to segment the index map.
For each binary segmented result, pixels could be divided into true positive (TP), false positive (FP), false negative (FN) and true negative (TN).
Precision of building detection is then the proportion of correctly detected building pixels in all detected building pixels. Recall is the the proportion of correctly detected building pixels in all building pixels.
They could be represented as below,
\begin{align}
Precision = \frac{TP}{TP+FP}, 
Recall = \frac{TP}{TP+FN}
\label{eq:precision-recall}
\end{align}

Let recall $r$ be the x axis and precision be the y axis, we could plot a precision-recall curve (Fig.\ref{fig:prcurve-Spacenet65}), where precision p(r) is a function of recall $r$. The average precision (\textit{AP}) is the area between the curve and the x axis.
Mean average precision is the mean value of \textit{AP} scores of images among one dataset.
\begin{align}
mAP = \frac{1}{n}\sum_{i=1}^{n}\int{P(r)dr}
\label{eq:map}
\end{align}

\textit{F-score} is defined as the harmonic mean of precision and recall. For an image, each segmented result corresponds to a \textit{F-score} and we choose the maximal \textit{F-score} as the final result. \textit{F-score} of a dataset is the average scores of all of the images in the dataset.
\begin{align}
F\textrm{-}score = \frac{2\times Precision\times Recall}{Precision+Recall}
\label{eq:f-score}
\end{align}

\begin{table*}[htb!]
	\caption{\label{table:map-f-datasets}\textit{F-score} and \textit{mAP} of several methods on three datasets. The best result is shown in \textbf{bold}.}
	\centering
	\begin{tabular}{m{3cm}<{\centering}|m{1.2cm}<{\centering}|m{1.5cm}<{\centering}|m{1.2cm}<{\centering}|m{1.5cm}<{\centering}|m{1.2cm}<{\centering}|m{1.5cm}<{\centering}}
		\hline
		\multirow{2}{*}{Methods}&
		\multicolumn{2}{c|}{Spacenet65}&\multicolumn{2}{c|}{Massachusetts}&\multicolumn{2}{c}{Potsdam}
		\cr\cline{2-7}&\textit{mAP}&\textit{F-score}&\textit{mAP}&\textit{F-score}&\textit{mAP}&\textit{F-score} \cr  
		\hline
		BASI & 0.34 & 0.44 & 0.32 & 0.40 & 0.34 & 0.44 \\
		MBI & 0.28 & 0.35 & 0.28 & 0.38 &  0.17 & 0.35 \\ 
		PBI & 0.27 & 0.37 & 0.25 & 0.36  & 0.41 & 0.50 \\ 
		HF-FCN & 0.04 & 0.12 & \textbf{0.76} & \textbf{0.74} & 0.03 & 0.10 \\
		GBI(ours) & \textbf{0.46} & \textbf{0.52} & 0.37 & 0.44 & \textbf{0.46} & \textbf{0.59} \\ \hline
	\end{tabular}
\end{table*}

\subsection{Results and analysis}

\begin{figure*}[htb!]
	\centering
	\includegraphics[height=.85\textheight]{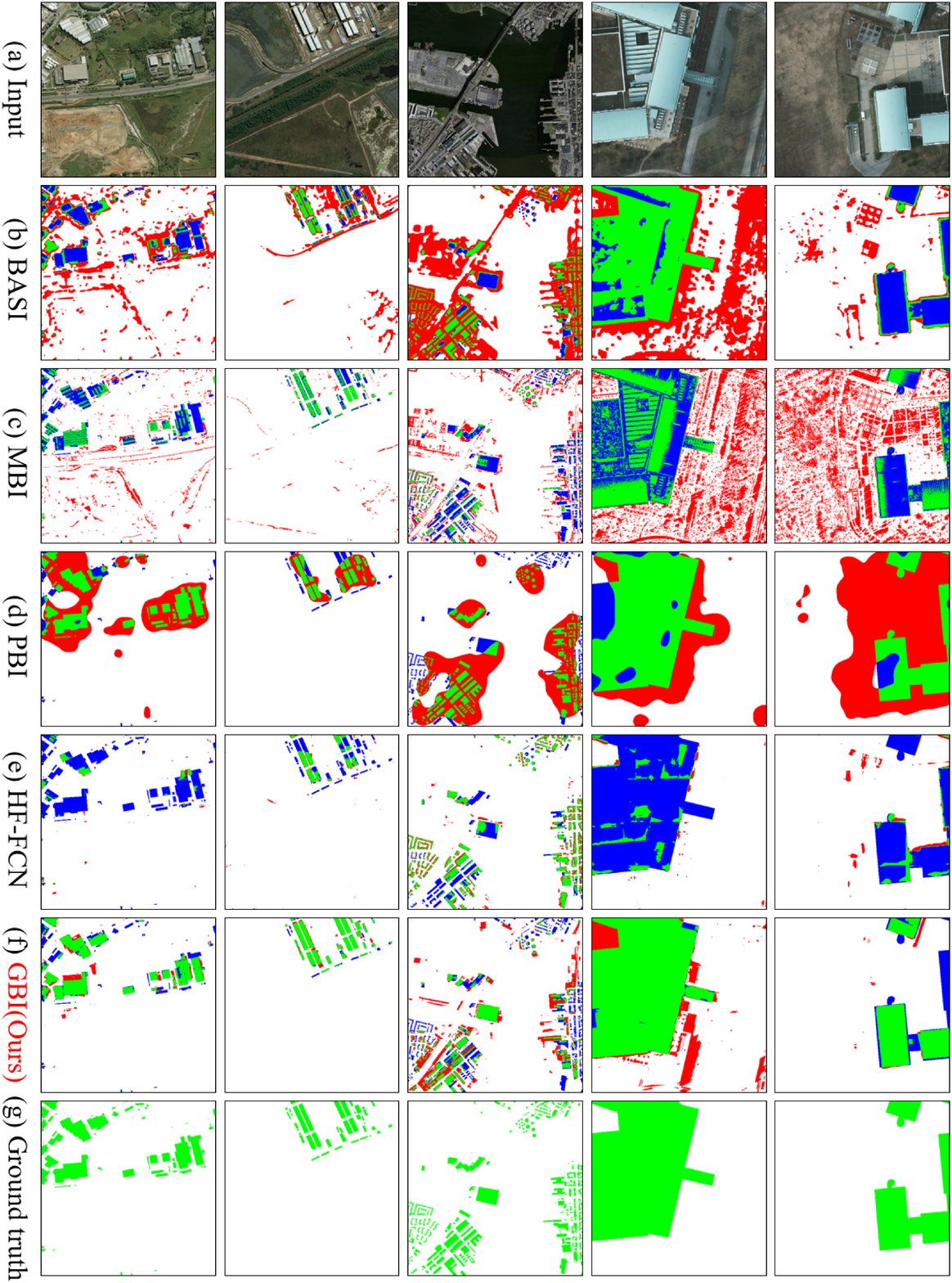}
	\caption{Segmented results of different building detection methods. Green areas are those correctly detected buildings and red areas are false detections. Blue areas represent the missed detected buildings. Compared with other methods, the result of GBI is very clean and much more similar to the ground truth. Furthermore, it preserves the geometric shape of single buildings.}
	\label{fig:segment_result}
\end{figure*}

The F-scores and mAP of all compared methods on three datasets are showed in the Table~\ref{table:map-f-datasets}.
It shows that our method achieved the best performance on both Spacenet-65 and Potsdam datasets.
On both two datasets, the \textit{mAP} score of GBI is higher than the second best method about nearly $15\%$ and its F-measure is higher than others' about $7\%$.
While on Massachusetts dataset, HF-FCN got the best performance, which is mainly due to the fact that the HF-FCN model is directly trained on this dataset.
However, HF-FCN showed very poor generalization capability when applied it to the other two datasets.
BASI, a texture-based method, also provided good performance on the three datasets although it was not the best one.
It generated many abundant false areas around buildings.
Like MBI, PBI also has such problem.
Its score is very high in Potsdam dataset but we could see from the segmented results (Fig.\ref{fig:segment_result}) that it only detected the built-in areas.
Those detected areas by BASI, MBI, and PBI are often correct but irregular, so they may be suited to act as a pre-processing step to find built-in areas before accurate building detection.

\begin{figure*}[htb!]
	\centering
	\includegraphics[width = .32 \linewidth]{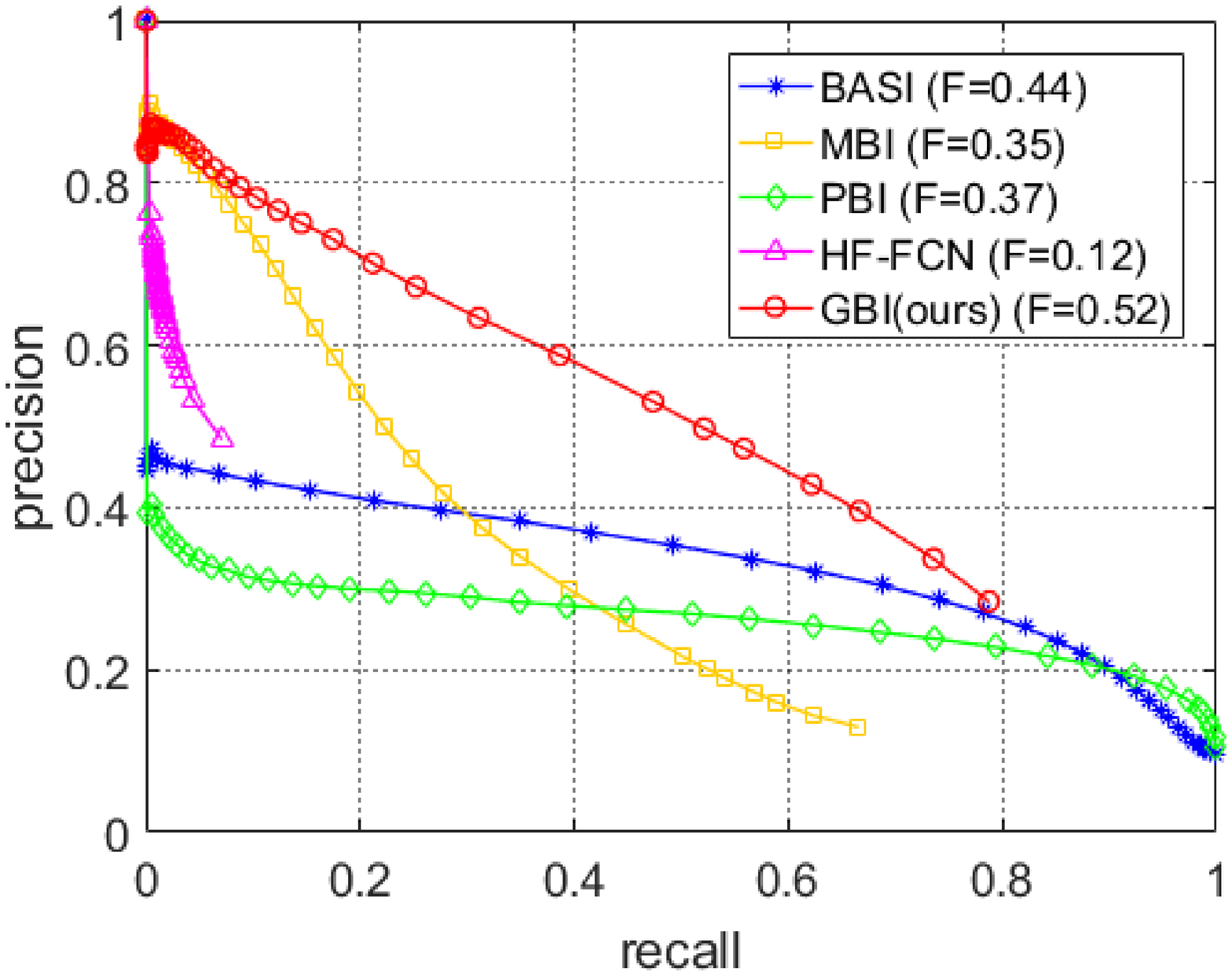}
	\includegraphics[width = .32 \linewidth]{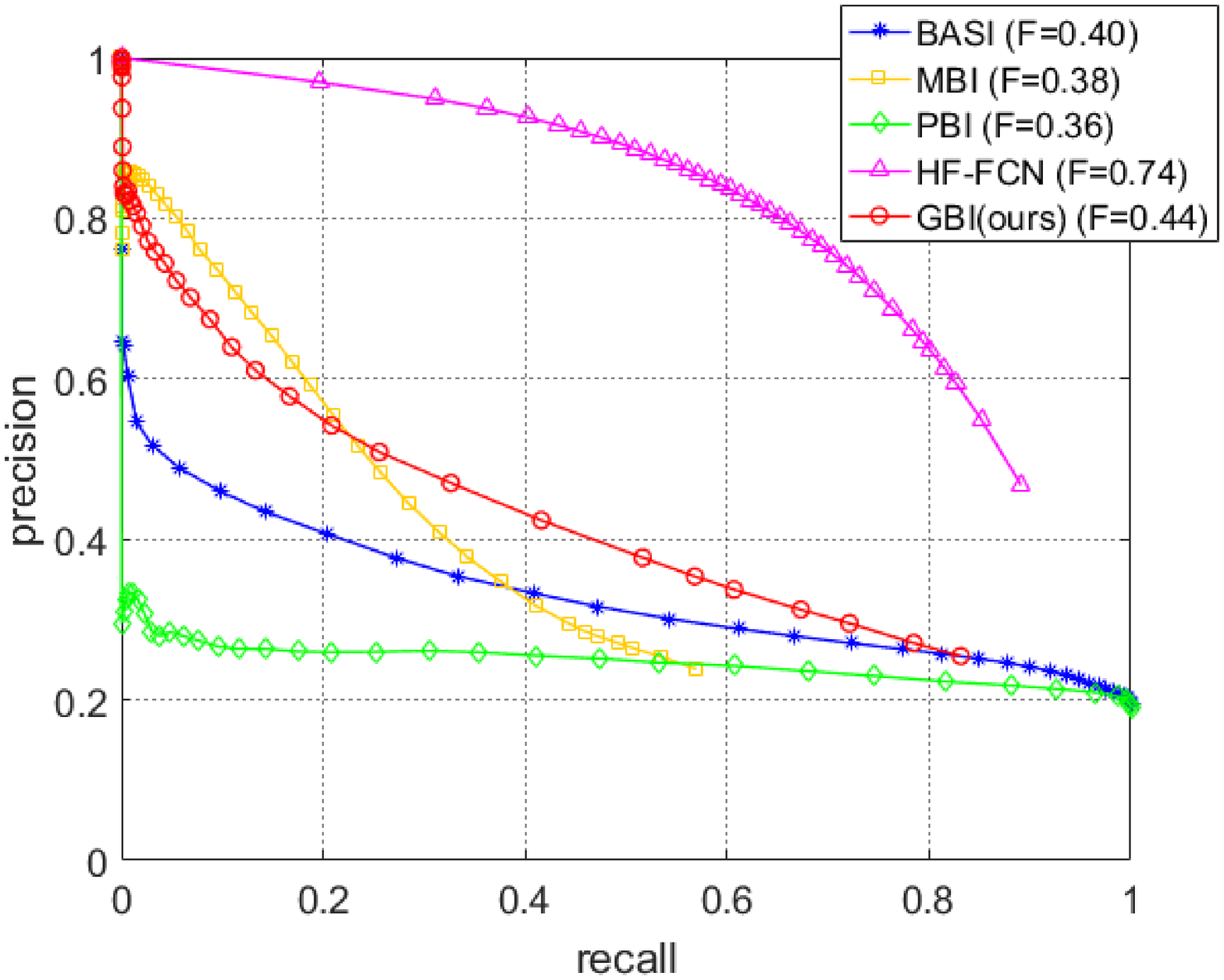}
	\includegraphics[width = .32 \linewidth]{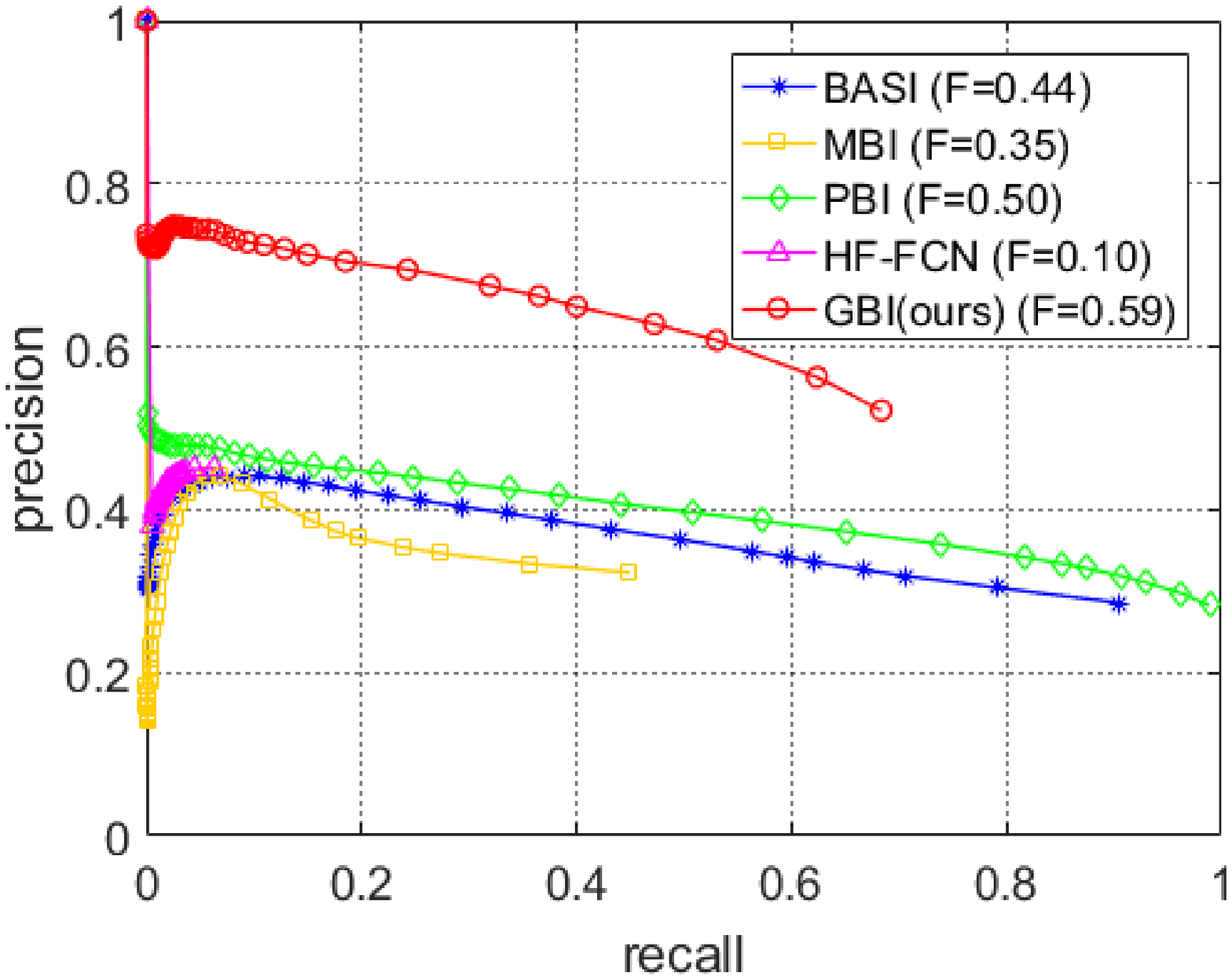}
	\caption{PR curves of different methods on Spacenet65, Massachusetts and Potsdam datasets respectively. There are five methods, and ours is showed in red. \textbf{F} indicates the F-score. 
	}
	\label{fig:prcurve-Spacenet65}
\end{figure*}

Fig.\ref{fig:prcurve-Spacenet65} illustrates the comparisons of precision-recall (PR) curves on the Spacenet65, Massachusetts and Potsdam datasets respectively.
From the results on Spacenet dataset, one can find that HF-FCN (the purple curve) has high precision but very low recall rate.
The curve of MBI drops very fast when the recall rate increases.
The detected buildings will be often incomplete and the geometric shape of buildings are broken.
Although BASI does not have such high precision, its precision only decrease a little when recall rate increases.
Compared with other methods, GBI shows a good performance. The curve of GBI (the red one) is always higher than others and achieves the highest F-score.
But we could also find that the recall rate of GBI is difficult to reach 1 because not every building could be detected by junctions and thus the geometric saliency is limited in those areas.
Besides GBI, other methods like HF-FCN and MBI also have such problems.

\begin{figure}[htb!]
	\centering
	\includegraphics[width = 0.45 \linewidth]{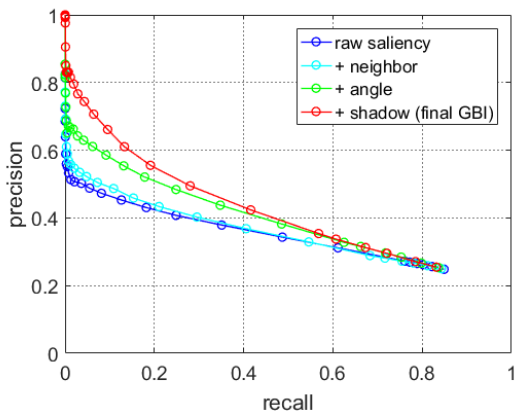}
	\caption{The PR curves of the proposed algorithms with different settings on Massachusetts dataset. }
	\label{fig:prcurve-param}
\end{figure}

\begin{figure*}[htb!]
	\centering
	\includegraphics[width=.81\linewidth]{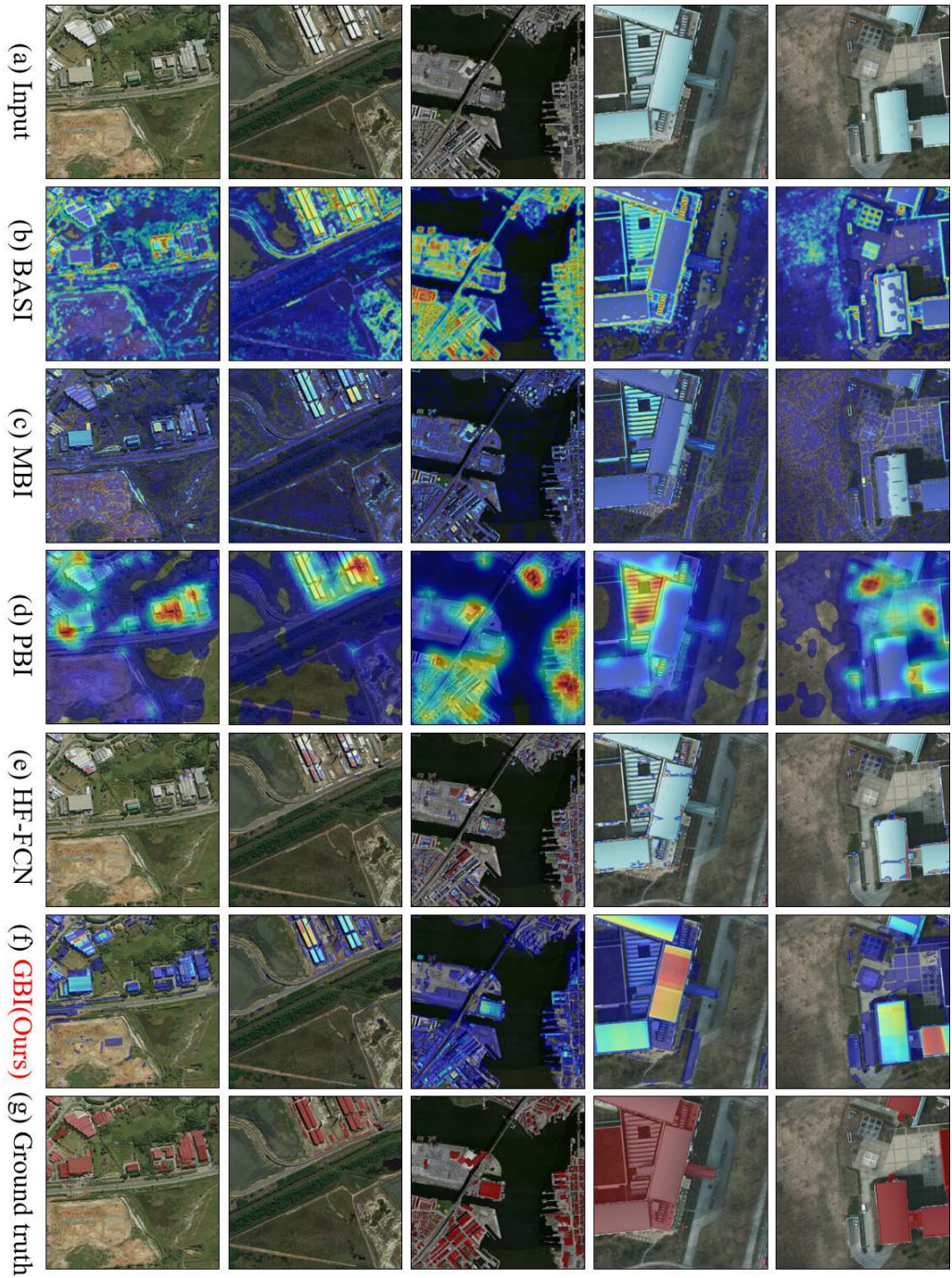}
	\vspace{-3mm}
	\caption{Building indexes generated from different methods. The color represent the value of building index, increase from blue to red. Results of BASI and PBI are distributed mainly around building areas and it is difficult to distinguish roads with single buildings. MBI misses some part of buildings and HF-FCN shows poor generality to different datasets. While building indexes generated by our proposed GBI are mainly distributed on single buildings and it is easy to find them out.}
	\label{fig:bindex}
\end{figure*}

Fig.~\ref{fig:bindex} and Fig.~\ref{fig:segment_result} display the corresponding building indexes and segmented results of 5 representative images in the three datasets.
Obviously, building indexes generated by GBI concentrate more on buildings. Thus the segmented result is very clean (with less false detections) and the precision is much higher than other methods.
While building indexes calculated by texture-based BASI are very mussy.
Besides the buildings, they also have large value on other non-building objects, like the long roads and textured forest.
The indexes given by MBI also suffer from such problems.
PBI only detects the building areas and it can not provide the accurate shape of buildings.
For HF-FCN, it has a high precision but the recall is very low.
In contrast, the results given by GBI preserve better the shape of single buildings.
For instance, from the segmented results, one can find that the detected buildings have a clear boundary.

Similar results can be found on the Potsdam dataset and Massachusetts dataset. 
It is worth noticing that HF-FCN shows almost perfect performance on Massachusetts dataset but extremely poor performance on the other two datasets. This is mainly due to the fact that the HF-FCN model was trained on Massachusetts dataset and there is no prior knowledge learned by the model from other two datasets. 
Thus, in the case that there is little prior knowledge about the data and no available annotated training samples, GBI can be the best choice for building extraction. 

\subsection{Ablation studies}

\begin{table*}[htb!]
	\caption{\label{table:map-f-param} Effect of taking different part of GBI into consideration. The best result is shown in \textbf{bold}.}
	\centering
	\begin{tabular}{m{5cm}<{\centering}|m{1.2cm}<{\centering}|m{1.5cm}<{\centering}|m{1.2cm}<{\centering}|m{1.5cm}<{\centering}|m{1.2cm}<{\centering}|m{1.5cm}<{\centering}}
		\hline
		\multirow{2}{*}{Methods}&
		\multicolumn{2}{c|}{Spacenet65}&\multicolumn{2}{c|}{Massachusetts}&\multicolumn{2}{c}{Potsdam}
		\cr\cline{2-7}&\textit{mAP}&\textit{F-score}&\textit{mAP}&\textit{F-score}&\textit{mAP}&\textit{F-score} \cr  
		\hline
		raw saliency & 0.41 & 0.48 & 0.31 & 0.41 & 0.45 & 0.58\\
		+ neighbor & 0.42 & 0.49 & 0.32 & 0.42 &  0.45 & \textbf{0.59}\\
		+ angle & 0.45 & \textbf{0.52} & 0.35 & \textbf{0.44}  & \textbf{0.46} & \textbf{0.59}\\
		+ shadow (final GBI) & \textbf{0.46} & \textbf{0.52} & \textbf{0.37} & \textbf{0.44} & \textbf{0.46} & \textbf{0.59}\\ \hline
	\end{tabular}
\end{table*}

In the computations of GBI, there are several parts, including the junction significance $\rho$ (raw saliency of single junction), neighboring information (pairwise geometric saliency), the distribution of junction's angle and the shadow information.
It is of great interest to study the effect of these different parts to the final accuracy of building detection.

In this experiment, we make some ablation studies by changing the way to generate geometric saliency. First, we call it the raw saliency when using only the significance of junctions and there is no pairwise term, {\em i.e.} the first-order saliency in Eq.\ref{eq:first-order} is changed to $(1- \rho_\jmath)$. We then add the pairwise term, the prior distribution of junction's angle, and the shadow information to the raw saliency one after another. Observe that the forth one with shadow information is the final GBI used in the previous experiments.
The result of this experiment is shown in Table.\ref{table:map-f-param} and the precision-recall curves evaluated on Massachusetts dataset is displayed in Fig.\ref{fig:prcurve-param}.

One can find that ``new-introduced" information have positive effect to the accuracy on Massachusetts and Spacenet dataset. More precisely, taking the distribution of junction's angle into consideration actually increases the performance largely. Massachusetts dataset contains mainly urban areas and thus buildings are more likely to have regular shapes. The introduction of angle's distribution will contribute to more salient junctions inside building areas. For the computation of GBI, we use the parallelograms of junctions.  When there are neighboring junctions located in a same building, their parallelograms will have overlapping areas. In such cases, the first-order geometric saliency indeed implicitly contains the neighboring information and the pairwise term will not help too much. In addition, shadow information also helps to improve the performances as Massachusetts dataset has many shadows inside it. However, the improvements on Potsdam dataset were not remarkable. In this dataset, buildings have very big size and there are only several buildings in an image due to the very high resolution. Meanwhile, shadows are also very little because of low lightness. In such cases, new information are hard to help improve accuracy. But the geometric structure is very salient in such high resolution, junctions are generally located around buildings and thus generates reasonable results. 
\section{Conclusion}
\label{sec:conclusion}
In this paper, we proposed a new method to calculate the building index to extract building in VHR-RS images, which is based on the defined geometric saliency of mid-level geometrical structures, {\em i.e.} junctions.
Our method achieves the state-of-the-art performances in contrast with existing methods which use the building index for extracting buildings from remote sensing images.
Furthermore, as the resolution and details of images increase, the performance of GBI grows obviously.
The building areas detected by GBI have a clearer boundary and less redundant cluttered areas than other methods and are much convenient to be applied in real application.


For further studies, in our current work, we only consider the geometric cues in panchromatic remote sensing images.
The combination of other cues like textures and spectral could also help.
For example, the NDVI could deal with the farmlands.
In the future, fusing different type of features will be a perspective way to improve the accuracy of building detection.

{\small
\bibliographystyle{apalike}
\bibliography{refs}
}

\end{document}